\documentclass[fleqn,10pt]{wlscirep}
\usepackage[utf8]{inputenc}
\usepackage[T1]{fontenc}
\usepackage{soul}
\usepackage{multirow}
\title{RGB-D-based Stair Detection using Deep Learning for Autonomous Stair Climbing}

\author[1]{Chen Wang}
\author[1,+]{Zhongcai Pei}
\author[1,+]{Shuang Qiu}
\author[1,*]{Zhiyong Tang}
\affil[1]{School of Automation Science and Electrical Engineering, Beihang University, Beijing, 100191, China}

\affil[*]{zyt\_76@buaa.edu.cn}

\affil[+]{these authors contributed equally to this work}

\begin{abstract}
Stairs are common building structures in urban environments, and stair detection is an important part of environment perception for autonomous mobile robots. Most existing algorithms have difficulty combining the visual information from binocular sensors effectively and ensuring reliable detection at night and in the case of extremely fuzzy visual clues. To solve these problems, we propose a neural network architecture with RGB and depth map inputs. Specifically, we design a selective module, which can make the network learn the complementary relationship between the RGB map and the depth map and effectively combine the information from the RGB map and the depth map in different scenes. In addition, we design a line clustering algorithm for the postprocessing of detection results, which can make full use of the detection results to obtain the geometric stair parameters. Experiments on our dataset show that our method can achieve better accuracy and recall compared with existing state-of-the-art deep learning methods, which are 5.64$\%$ and 7.97$\%$, respectively, and our method also has extremely fast detection speed. A lightweight version can achieve 300 + frames per second with the same resolution, which can meet the needs of most real-time detection scenes. 
\end{abstract}
\begin{document}

\flushbottom
\maketitle
%
%
\thispagestyle{empty}

\section*{Introduction}

With a long research history for autonomous mobile robots, stair climbing is a fundamental problem and has a wide range of applications. The premise of climbing stairs is to obtain the geometric parameters of stairs, so the environmental perception system of an autonomous mobile robot must have the ability to detect stairs. For stair detection, there are two kinds of methods according to the method of stair feature extraction: line extraction methods and plane extraction methods.

Line extraction methods abstract the features of stairs as a collection of continuous lines in an image and use relevant algorithms to extract the features of lines in RGB or depth maps. There are two kinds of mainstream algorithms for feature extraction: traditional image processing algorithms and deep learning algorithms. The former use Canny edge detection, a Hough transform and other methods to extract lines\cite{bib1, bib2, bib3, bib4}. The latter use deep learning computer vision to make a network extract line features through learning on a dataset with annotations\cite{bib5}. The main idea of plane extraction methods is to detect a group of continuously distributed parallel planes in point clouds. Point clouds are divided into planes using a plane segmentation algorithm, and then the stair planes can be extracted according to some manually designed rules\cite{bib6, bib7, bib8}. In addition, some methods first use deep learning to locate a region of interest(ROI) containing the stairs and then use a traditional image processing algorithm for line extraction within the ROI\cite{bib9}. These methods often have poor real-time performance because of the two detection stages.

According to the information input mode for a detection algorithm, stair detection can also be divided into monocular and binocular detection methods. Monocular detection methods extract features only from single-mode perceptual information. Some common workflows are as follows. Line extraction methods are applied to obtain stair features in RGB maps, and the corresponding key points are obtained from the point clouds. Then, the geometric stair parameters are obtained\cite{bib10}. Line extraction methods can also be applied to obtain the stair features in depth maps to obtain the geometric stair parameters\cite{bib11}. Plane segmentation methods are applied to obtain the stair planes in point clouds to calculate the geometric stair parameters\cite{bib6, bib7, bib8}. The sensors used in monocular detection methods may be binocular, and only the single-mode information is used for feature extraction. Binocular detection methods extract features from multimodal perceptual information. Some common workflows are as follows. After using line extraction methods to  extract features from RGB and depth maps, the features from different modes are fused artificially using certain rules\cite{bib12, bib13}. Monocular detection methods often have better real-time performance but lower accuracy than binocular detection methods.

The above methods have long provided stair detection abilities for robots used in urban environments. However, there are still some challenging problems to be addressed. For line extraction methods, the edges of stairs in an RGB map are clear when ascending stairs but fuzzy when descending, which is the opposite in a depth map, namely, the RGB and depth map information are complementary to some extent. Therefore, monocular detection methods alone cannot obtain complete and reliable visual clues. For binocular detection methods, the detection results from the RGB and depth maps need to be fused. Existing fusion methods are mainly based on artificially designed rules, which often do not have wide applicability. For plane extraction methods, algorithms that directly use plane segmentation algorithms to process the point clouds returned from a depth sensor often have poor real-time performance, which is caused by the very large amount of point cloud data.

\begin{figure}[ht]
\centering
\includegraphics[width=0.8\linewidth]{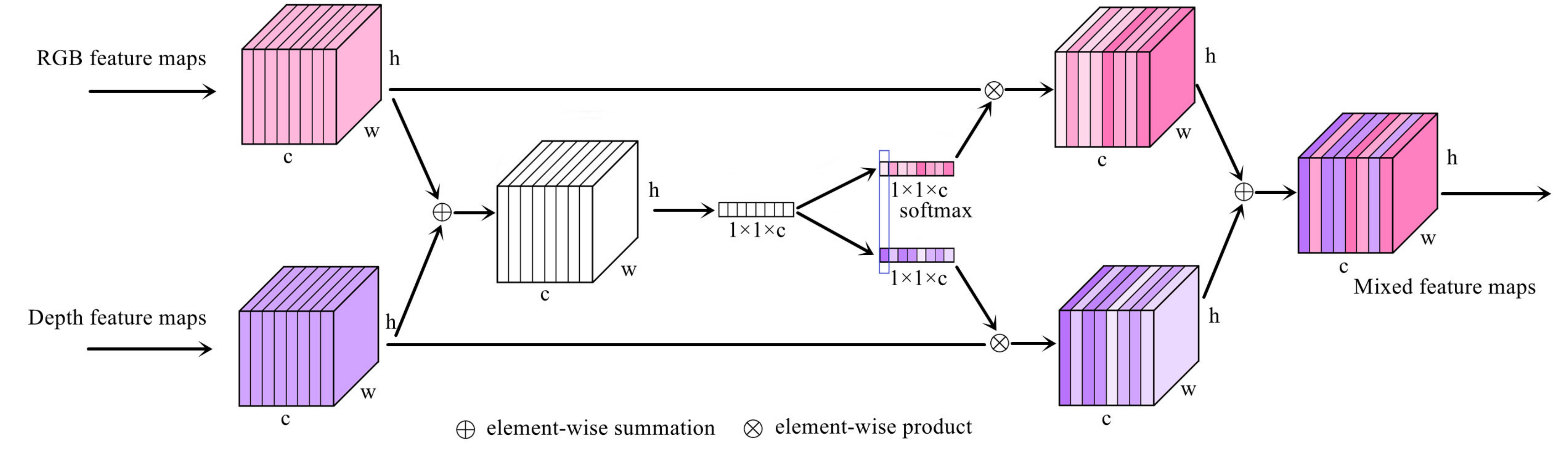}
\caption{The selective module. First, the feature maps containing RGB information and the feature maps containing depth information are added element by element. Then, global average pooling is applied to obtain a 1 $\times$ 1 $\times$ c vector as the feature descriptor of the RGB or depth feature maps, and the softmax activation function is applied to the channel c dimension to obtain two feature descriptors that restrict each other. Finally, the feature descriptor is multiplied by the corresponding original feature maps element by element, and the depth and RGB feature maps are added to obtain the mixed feature maps.}
\label{fig1}
\end{figure}

In previous works, among methods based on line extraction, StairNet\cite{bib5} has achieved breakthrough accuracy and extremely fast speed under the condition of monocular vision by virtue of its novel stair feature representation method and end-to-end CNN architecture. However, due to the limitations of monocular vision, the method has poor performance in night scenes with extreme lighting, scenes with extremely fuzzy visual clues and scenes with a large number of objects similar to stairs. This study adds depth map input into the network on the basis of StairNet. Through a selective module, the network can learn the complementary relationship between the RGB map and the depth map and effectively combine the information from both in different scenes, as shown in Fig. \ref {fig1}.

In summary, the contribution of this work can be summarized in three parts: 1) Based on StairNet, we add depth map input into the network, and through a selective module, the network can learn the complementary relationship between the RGB map and the depth map so that a feature fusion process can be integrated into the neural network and eliminate the artificially designed rules. We call this network StairNetV2. Compared with previous methods, we can obtain extremely fast detection speed while processing multimodal perception information and still have excellent performance in extreme lighting scenes at night, scenes with extremely fuzzy visual cues, and scenes with a large number of objects similar to stairs. 2) We design a line clustering algorithm based on the least squares method, which can quickly cluster the line segments belonging to the same line in an image. The algorithm can make full use of the detection results to obtain the geometric stair parameters. 3) We provide an RGB-D dataset with fine annotations for stair detection research. The training set contains 4776 images (including 2388 RGB images and the corresponding 2388 depth images), and the validation set contains 1216 images (including 608 RGB images and the corresponding 608 depth images). Each label contains the locations of the two endpoints and the classification(convex/concave) of each stair line.

\section*{Related works}

The essence of the RGB-D multimode fusion stair detection methods is an RGB-D multimode fusion object detection method. In this section, we briefly introduce some RGB-D-based object detection and semantic segmentation methods, as well as some RGB-D-based stair detection methods.

\subsection*{Object detection and semantic segmentation methods based on RGB-D}

Object detection and semantic segmentation are fundamental and long-standing problem in computer vision. With the development of low-cost depth sensors, there are an increasing number of object detection methods and semantic segmentation methods based on RGB-D. However, there is no established methodology to perfectly fuse these two modalities inside a CNN. Common methods include the following: 1) Take the depth map as an additional input branch and fuse it with the RGB branch. The main disadvantage is that processing two branches at the same time requires additional computation. 2) Take the depth map as an operation. 3) Take the depth map as a prediction. This method uses only the depth map in the training step so that the network can predict the depth, namely, depth estimation. This method does not require binocular sensors during inference, but the accuracy of the depth map is often worse\cite{bib14}. Considering the accuracy of geometric stair parameters, we focus on the first two methods.

For the first method, references\cite{bib15, bib16} directly adjust the three-channel RGB input to the four-channel RGB-D input, and the depth information is send to the network as the fourth input channel. References\cite{bib17, bib18} design two parallel branches to process RGB map and the depth map and then concatenate the feature vectors obtained from each branch to fuse the features. Finally, the number of channels is adjusted, and the softmax function is applied to obtain the final classification result. This simple way to fuse RGB and depth information in the input or output can not make full use of the complementary relationship between the RGB map and the depth map. To solve this problem, reference\cite{bib19} proposes canonical correlation analysis(CCA) to fuse the two kinds of features. Reference\cite{bib20} applies ensemble learning to combine an RGB map model and a depth map model. Reference\cite{bib21} proposes a multi-modal layer to combine the information from the RGB stream and the depth stream. Through multimodal learning, this layer can not only discover the most discriminative features for each modality but also harness the complementary relationship between them. Reference\cite{bib22} first trains an RGB network and a deep network separately and then deletes their softmax classification layers after training. Finally, a fusion network with the softmax function is trained to obtain the final classification. For the second method, Reference\cite{bib23} proposes a depth-similar convolution operation, which converts the unique pixel relationship in the depth map into depth-similar weight, dots with the corresponding convolution kernel, and convolves with the input feature map for the final output feature map. 

\subsection*{Stair detection methods based on RGB-D}

Most stair detection methods based on RGB-D are limited by the vision algorithm used, and their combination of RGB and depth features is often rigid. In reference\cite{bib12}, the RGB and depth maps are obtained using a depth camera, and then prior knowledge is used to determine which map is used for detection. Finally, stair lines are obtained through edge and line detection algorithms. In reference\cite{bib13}, the RGB and depth maps are processed at the same time and the edge information in the maps is obtained. Then the local binary pattern(LBP) feature is extracted from the RGB map, and the one-dimensional depth feature is extracted from the depth map. Finally, the features are combined and classified using a  support vector machine(SVM). Reference\cite{bib2} first applies a Sobel operator to extract the edges from the RGB map, and then a Hough transform is applied to extract the straight lines from the edge image. Finally, one-dimensional depth features are extracted from the depth map to distinguish stairs from objects with textures similar to stairs, such as pedestrian crosswalks. Reference\cite{bib24} develops an RGB-D stair recognition system that applies unsupervised domain adaptation learning to help visually impaired people navigate independently in unfamiliar environments. In this study, the three-channel RGB input is adjusted to a four-channel RGB-D input to fuse depth information.

\section*{Methods}

In this section, we describe the details of our method from four aspects, including the StairNetV2 network architecture, the loss function with dynamic weights, stair line clustering based on the least squares method and the method to obtain the geometric stair parameters based on attitude angles.

\subsection*{Network architecture}

\begin{figure}[ht]
\centering
\includegraphics[width=0.8\linewidth]{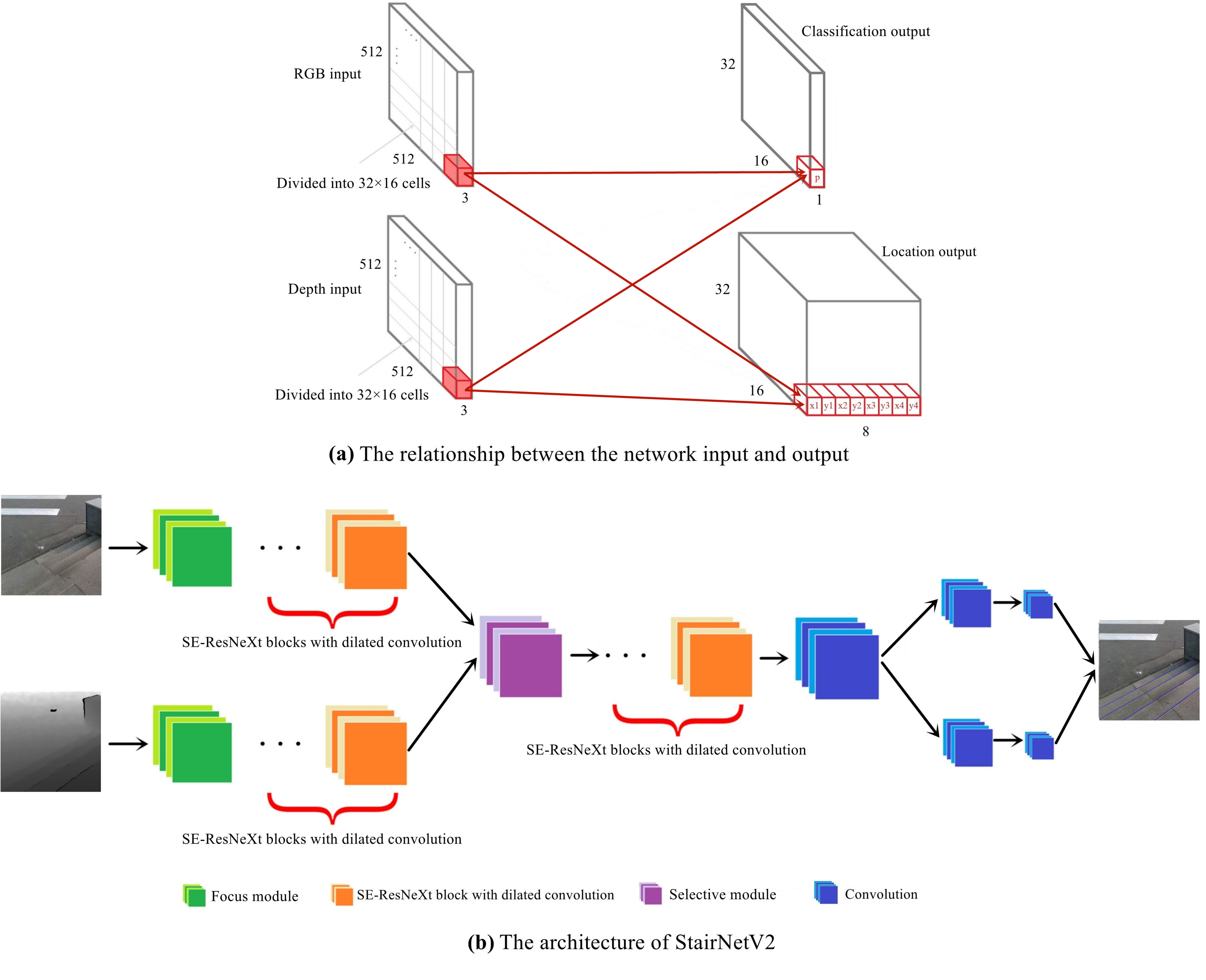}
\caption{Illustration of the network architecture. Figure (a) shows the relationship between the network input and output. The network takes a 512 $\times$ 512 RGB map and a 512 $\times$ 512 depth map as input, and the output is divided into two branches. The size of the output tensor of the classification branch is 32 $\times$ 16 $\times$ 1, which is used to judge whether a cell contains stair lines. The size of the output tensor of the location branch is 32 $\times$ 16 $\times$ 8, which is used to predict the locations of two sets of stair lines. Figure (b) shows the architecture of our network. The backbone contains two independent input branches with a focus module and several SE-ResNeXt blocks with dilated convolution. The selective module is applied to fuse features. Then, the SE-ResNeXt blocks are applied for further feature extraction.}
\label{fig2}
\end{figure}

As described above, we add depth map input to the network based on StairNet and fuse the features from the RGB map and the depth map using a selective module. On this basis, to accelerate postprocessing of the network output results, we adjust the size of the final output feature map to 32 $\times$ 16. To make the classification result more robust, we adjust the previous 3 classifications to 2 classifications; that is, we no longer judge whether a stair line is a concave line or a convex line, but only judge whether each cell contains stair lines. This means that we only distinguish between the foreground and background, and the judgment of concave and convex lines can be easily obtained through prior knowledge. In summary, as shown in Fig. \ref {fig2}, our model takes a 512 $\times$ 512 full-color image and a 512 $\times$ 512 depth image as input and processes them with a fully convolutional architecture. A feature map with a size of 32 $\times$ 16 is obtained after four downsampling operations. The output of the network is divided into two branches, and each cell location in the 3D output tensor is associated with a multidimensional vector. The output tensor obtained from the classification branch has a size of  32 $\times$ 16 $\times$ 1, which is used to determine whether a cell contains stair lines. The output tensor obtained from the location branch has a size of 32 $\times$ 16 $\times$ 8, and each cell predicts two sets of locations (x1, y1, x2, y2) and (x3, y3, x4, y4). Regardless of the posture of the stair line in the cell, (x1, y1) and (x3, y3) always represent the locations of the left endpoints, and (x2, y2) and (x4, y4) always represent the locations of the right endpoints. In addition, because a few cells will contain two stair lines, we  predict two sets of locations. For cells containing only one stair line, the two sets of locations have the same values.

Compared with the StairNet architecture, the receptive field is further expanded due to the increase in downsampling times, so atrous spatial pyramid pooling (ASPP)\cite{bib25} is removed. StairNet applies asymmetric hole convolution to take advantage of the prior knowledge that stair lines are usually distributed horizontally in an image. In StairnetV2, we take advantage of this prior knowledge from the perspective of the loss function, and the asymmetric hole convolution in squeeze-and-excitation(SE)-ResNeXt\cite{bib26, bib27} block is replaced with the standard hole convolution. In addition, we retain the focus module\cite{bib28} as the initial downsampling of the network. See Table \ref{tab1} for the detailed StairNetV2 architecture.

\begin{table}[ht]
\centering
\begin{tabular}{|l|l|l|l|l|l|}
\hline
\multicolumn{2}{|l|}{Name} & \multicolumn{2}{|l|}{Type} &\multicolumn{2}{|l|}{Output size}\\
\hline
RGB branch & Depth branch & RGB branch & Depth branch & RGB branch & Depth branch \\
\hline
Initial & Initial & Tensor slice & Tensor slice & 256 $\times$ 256 $\times$ 32 & 256 $\times$ 256 $\times$ 32\\
\hline
Bottleneck 1.0 & Bottleneck 1.0 & Downsampling & Downsampling & 128 $\times$ 128 $\times$ 128 & 128 $\times$ 128 $\times$ 128\\
\hline
Bottleneck 1.1 & Bottleneck 1.1 & & & 128 $\times$ 128 $\times$ 128 & 128 $\times$ 128 $\times$ 128\\
\hline
Bottleneck 1.2 & Bottleneck 1.2 & & & 128 $\times$ 128 $\times$ 128 & 128 $\times$ 128 $\times$ 128\\
\hline
\multicolumn{2}{|l|}{Seletive module} & \multicolumn{2}{|l|}{} &\multicolumn{2}{|l|}{128 $\times$ 128 $\times$ 128}\\
\hline
\multicolumn{2}{|l|}{Bottleneck 2.0} & \multicolumn{2}{|l|}{Downsampling} &\multicolumn{2}{|l|}{64 $\times$ 64 $\times$ 256}\\
\hline
\multicolumn{2}{|l|}{Bottleneck 2.1} & \multicolumn{2}{|l|}{Dilated 2} &\multicolumn{2}{|l|}{64 $\times$ 64 $\times$ 256}\\
\hline
\multicolumn{2}{|l|}{Bottleneck 2.2} & \multicolumn{2}{|l|}{} &\multicolumn{2}{|l|}{64 $\times$ 64 $\times$ 256}\\
\hline
\multicolumn{2}{|l|}{Bottleneck 2.3} & \multicolumn{2}{|l|}{Dilated 4} &\multicolumn{2}{|l|}{64 $\times$ 64 $\times$ 256}\\
\hline
\multicolumn{2}{|l|}{Bottleneck 2.4} & \multicolumn{2}{|l|}{} &\multicolumn{2}{|l|}{64 $\times$ 64 $\times$ 256}\\
\hline
\multicolumn{2}{|l|}{Bottleneck 2.5} & \multicolumn{2}{|l|}{Dilated 8} &\multicolumn{2}{|l|}{64 $\times$ 64 $\times$ 256}\\
\hline
\multicolumn{2}{|l|}{Bottleneck 2.6} & \multicolumn{2}{|l|}{} &\multicolumn{2}{|l|}{64 $\times$ 64 $\times$ 256}\\
\hline
\multicolumn{2}{|l|}{Bottleneck 2.7} & \multicolumn{2}{|l|}{Dilated 16} &\multicolumn{2}{|l|}{64 $\times$ 64 $\times$ 256}\\
\hline
\multicolumn{4}{|l|}{Repeat bottlenecks 2.0 to 2.7} & \multicolumn{2}{|l|}{32 $\times$ 32 $\times$ 256}\\
\hline
\multicolumn{2}{|l|}{Conv 3 $\times$ 3} & \multicolumn{2}{|l|}{Down sampling with stride=(1,2)} &\multicolumn{2}{|l|}{32 $\times$ 16 $\times$ 128}\\
\hline
classification   & location   & classification  & location   & classification   & location  \\
\hline
Conv 3 $\times$ 3   & Conv 3 $\times$ 3   &   &    & 32 $\times$ 16 $\times$ 128   & 32 $\times$ 16 $\times$ 128  \\
\hline
Conv 1 $\times$ 1   & Conv 1 $\times$ 1   &   &    & 32 $\times$ 16 $\times$ 1   & 32 $\times$ 16 $\times$ 8  \\
\hline
& Sigmoid   &   &  Activation  & 32 $\times$ 16 $\times$ 1   & 32 $\times$ 16 $\times$ 8  \\
\hline
\end{tabular}
\caption{\label{tab1}StairNetV2 architecture with an input size of 512 $\times$ 512. At the end of the location branch, we add the sigmoid activation function to limit the output to (0,1) to accelerate network convergence and obtain the normalized coordinates of the stair line in each cell.}
\end{table}

\subsection*{Loss function with dynamic weights}\label{subsec4_2}

In StairNet, based on the multitask loss idea of an object detection task, the loss function is as follows:

\begin{equation}
	\begin{split}
		L(\{p_{ij}\},\{x_{ij}\}, \{y_{ij}\})=\frac{1}{MN}\sum_{i}^{M}\sum_{j}^{N}(L_{cls}(p_{ij},p_{ij}^*)+\lambda (p_{ij}L_{loc}(x_{ij},x_{ij}^*) + \alpha p_{ij}L_{loc}(y_{ij},y_{ij}^*)))\label{eq1}
	\end{split}
\end{equation}

Where, $ L_{cls} $ and $ L_{loc} $ represent the classification loss and location loss, respectively, and $ L_{loc} $ is divided into abscissa loss and ordinate loss. M and N represent the number of cells in a cloumn and a row, namely, 32 $\times$ 16, and i and j represent the position of the cell in the image. $ p_{ij} $ is a one-dimensional vector that indicates the probability that the cell contains stair lines, and its corresponding ground truth is $ p_{ij}^* $. $ x_{ij} $ is a 4-dimensional vector that represents the 4 abscissa coordinates predicted in the cell, and its corresponding ground truth is $ x_{ij}^* $. $ y_{ij} $ is a 4-dimensional vector that represents the 4 ordinate coordinates predicted in the cell, and its corresponding ground truth is $ y_{ij}^* $. $ \lambda $ and $ \alpha $ are the weight coefficients, $ \lambda $ is used to allocate the weight between the classification loss and the location loss, and $ \alpha $ is used to allocate the weight between the abscissa location loss and the ordinate location loss. In StairNet, $ \lambda $ and $ \alpha $ are set to 4.

This method of setting the weights to a fixed value provides a direction for the learning and convergence of the network to a certain extent. However, the values of weights should be different at different stages of network convergence. Therefore, we propose a loss function with dynamic weights, and equation \ref{eq1} is rewritten as:

\begin{equation}
	\begin{split}
		L(\{p_{ij}\},\{x_{ij}\}, \{y_{ij}\})=\frac{1}{MN}\sum_{i}^{M}\sum_{j}^{N}(L_{cls}(p_{ij},p_{ij}^*)+\alpha p_{ij}L_{loc}(x_{ij},x_{ij}^*) + \beta p_{ij}L_{loc}(y_{ij},y_{ij}^*))\label{eq2}
	\end{split}
\end{equation}

Where, $\alpha$ and $\beta$ represent the weight of the abscissa location loss and the weight of the ordinate location loss, respectively. The remaining parameters are the same as in equation \ref{eq1}. At the beginning of the training, $\alpha$ and $\beta$ are set to 10. After each epoch, $\alpha$ and $\beta$ are adjusted dynamically according to the evaluation results on the validation set. The evaluation method investigates the total error of the abscissa and ordinate prediction location of the positive class on the validation set, which are recorded as $ X_{error} $ and $ Y_{error} $ respectively. The corresponding formula is as follows:

\begin{equation}
	\begin{split}
		X_{error}=\sum_{k=0}^{V}\sum_{i}^{M}\sum_{j}^{N}p_{ij}^*E_{loc}(x_{ij},x_{ij}^*)\label{eq3}
	\end{split}
\end{equation}

\begin{equation}
	\begin{split}
		Y_{error}=\sum_{k=0}^{V}\sum_{i}^{M}\sum_{j}^{N}p_{ij}^*E_{loc}(y_{ij},y_{ij}^*)\label{eq4}
	\end{split}
\end{equation}

Where V represents the number of samples on the validation set, and i, j, M, N, $ p_{ij}^* $, $ x_{ij} $, $ x_{ij}^* $, $ y_{ij} $ and $ y_{ij}^* $ rhave the same meanings as equation \ref{eq1}. $ E_{loc} $ represents the error measurement method, and the L1-norm is used here. After obtaining $ X_{error} $ and $ Y_{error} $, we use equations \ref{eq5} and \ref{eq6} to adjust $ \alpha $ and $ \beta $. It should be noted that if the adjusted $ \alpha $ and $ \beta $ are less than threshold $ \sigma $ (set as 0.5 in the study), they will not be adjusted.

\begin{equation}
	\begin{split}
		\alpha_{i}=\alpha_{i-1}+\frac{X_{error}-Y_{error}}{Max(X_{error},Y_{error})}\label{eq5}
	\end{split}
\end{equation}

\begin{equation}
	\begin{split}
		\beta_{i}=\beta_{i-1}-\frac{X_{error}-Y_{error}}{Max(X_{error},Y_{error})}\label{eq6}
	\end{split}
\end{equation}

Where i represents the current epoch, i-1 represents the previous epoch, and i>=1. $ Max(X_{error},Y_{error}) $ represents the larger of $ X_{error} $ and $ Y_{error} $.

We apply the loss function with dynamic weights to enable the network to adjust its focus on the horizontal and vertical coordinate location loss. From the perspective of the loss function, we use the prior knowledge that stair lines are usually distributed horizontally in an image. Compared with the method in StairNet that uses prior knowledge by designing a module with a special structure, our method has better flexibility and wider applicability.

\subsection*{Stair line clustering}\label{subsec4_3}

\begin{figure}[ht]
\centering
\includegraphics[width=0.4\linewidth]{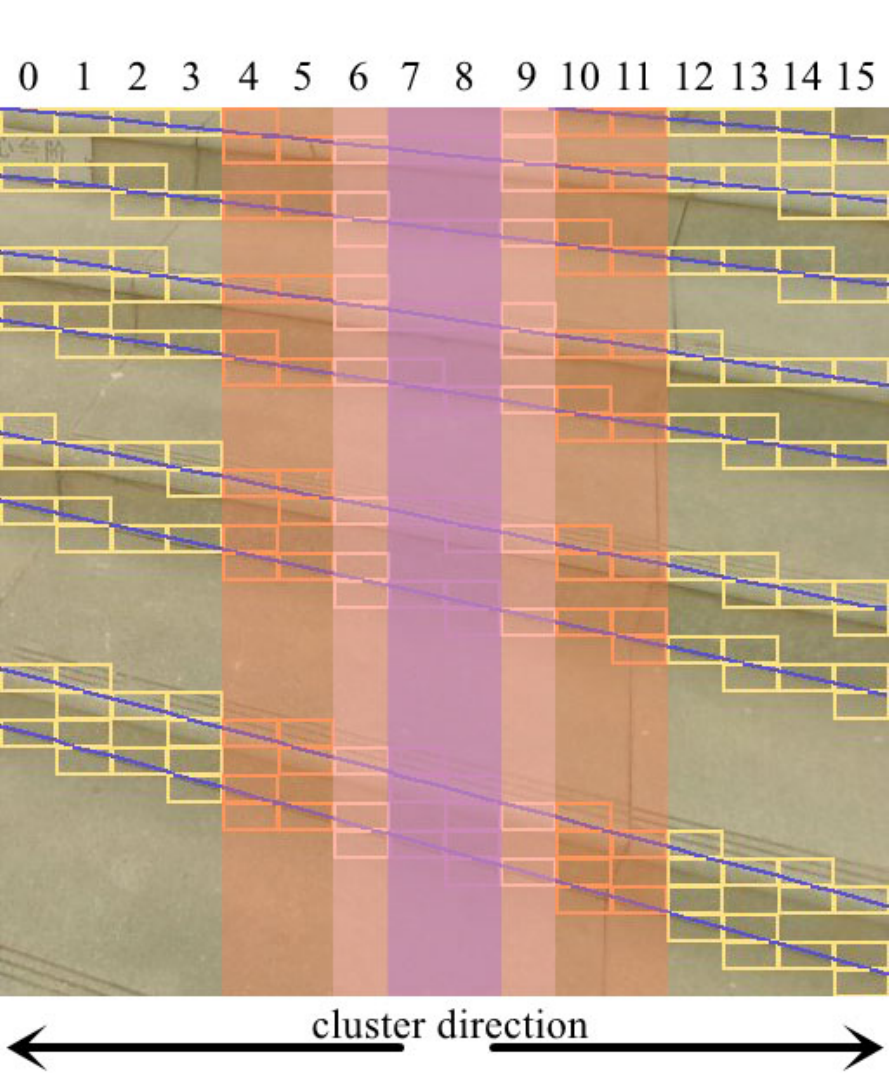}
\caption{Illustration of stair line clustering based on the least squares method. Purple, pink, orange and yellow represent the relevant areas of each cluster. First, the initial slope k and intercept b are calculated using the least squares method according the coordinates in the middle two columns of cells. Then the remaining cells are clustered from the middle to both sides to determine which line each cell covers or add new lines to the results. Finally, k and b are recalculated, and the process is repeated until clustering is completed.}
\label{fig3}
\end{figure}

As described above, using StairNetV2, the stair information we finally obtained is stored in two tensors. To finally obtain the geometric stair parameters, we need to group the results of these tensors, namely, to cluster the cells belonging to the same stair lines. Based on the least squares method, we design a stair line clustering method, as shown in Fig. \ref{fig3}. First, the initial slope k and intercept b are calculated using the least squares method according to the coordinates in the cells of the middle two columns, namely, column 7 and column 8. Then, the remaining cells are clustered from the middle to both sides to determine which line each cell covers. If they are on existing lines, they are added to the corresponding set of lines. Otherwise, the new lines are added to the results. Finally, k and b are recalculated using the least squares method, and the process is repeated until clustering is completed.

We select the middle two columns as the initial set of lines because the sensor is often facing the stairs in the actual scene, and there are more stair lines in the middle area, the prediction results are more accurate because of more visual information than the side area. In addition, for the two sets of stair lines (x1, y1, x2, y2) and (x3, y3, x4, y4) predicted by each cell, we only take the former when they are very close.

\subsection*{Acquisition of geometric stair parameters}

\begin{figure}[ht]
\centering
\includegraphics[width=0.6\linewidth]{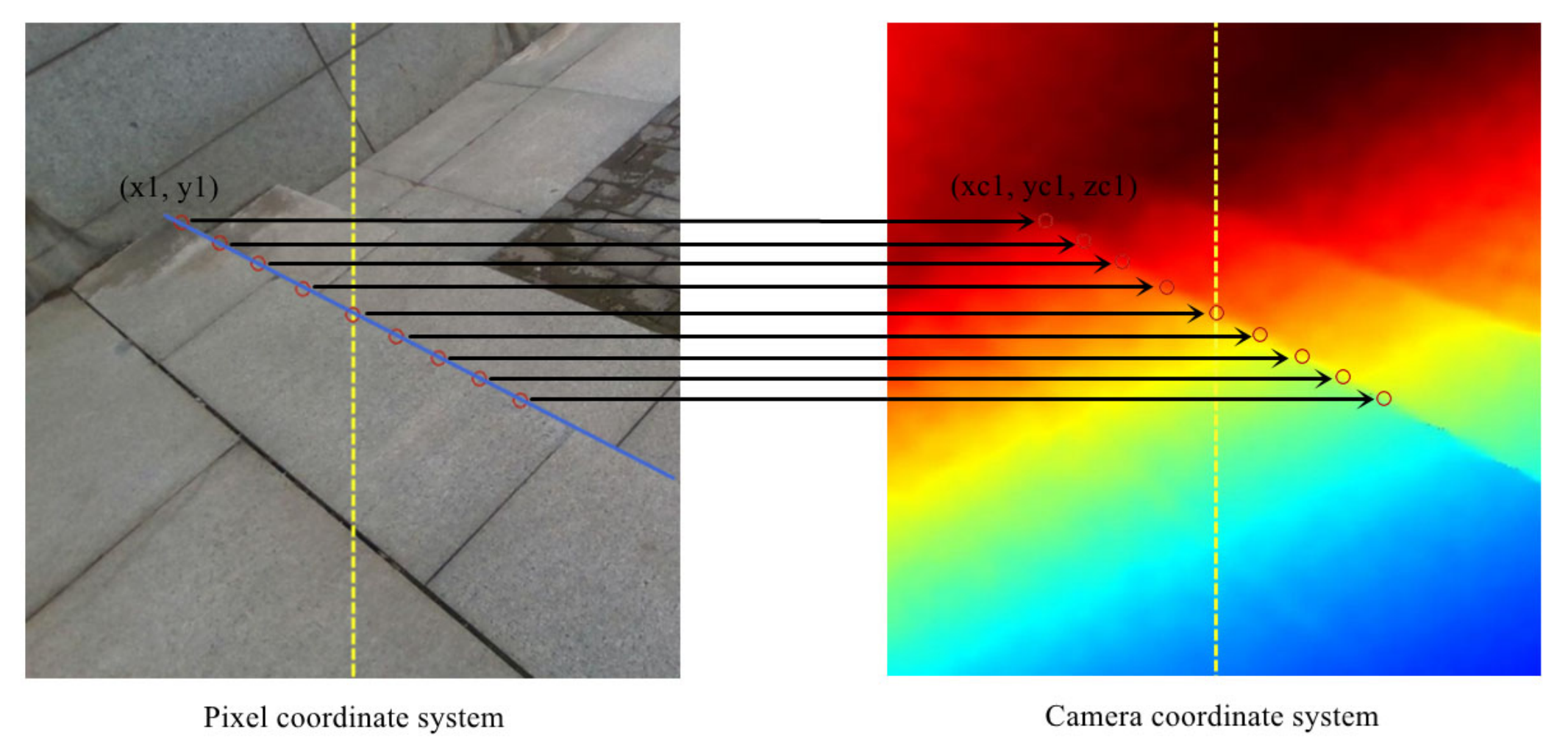}
\caption{Illustration of obtaining the stair sampling points. Taking the left endpoint of the stair line as an example, its coordinates in the pixel coordinate system are (x1, y1). Using the aligned point cloud data, its corresponding coordinates (xc1, yc1, zc1) in the camera coordinate system are obtained and the remaining sampling points are the same.}
\label{fig4}
\end{figure}

In section \href{subsec4_3}{"Stair line clustering"}, we sort the output results of the neural network through stair line clustering based on the least squares method and finally obtain the point sets and the corresponding slope and intercept describing each stair line. Since the slope and intercept reflect the information of all point sets, we use the six-tuple data (x1, y1, x2, y2, k, b) to describe a stair line, where x1, y1 and x2, y2 represent the left and right endpoints of the stair line, respectively, and k and b represent the slope and intercept, respectively. To make full use of the location information of the fitted stair line, we take 4 sampling points at symmetrical and equal intervals on the left and right sides of the image middle axis so we can obtain 9 sampling points. Next, we obtain the three-dimensional coordinates of the 9 sampling points in the camera coordinate system using the aligned point cloud data, as shown in Fig. \ref{fig4}.

After obtaining the three-dimensional coordinates of 9 sampling points in the camera coordinate system, we fit them to obtain a final set of three-dimensional coordinates to calculate the geometric stair parameters. We apply the least squares method of lines in three-dimensional space\cite{bib29} to fit these 9 sampling points, and the spatial straight line equation can be expressed as the intersecting line of two planes, as shown in equation \ref{eq7}:

\begin{equation}
	\begin{cases}
	
	x=k_{1}z+b_{1}\\
	y=k_{2}z+b_{2}
	
	\end{cases}\label{eq7}
\end{equation}

\begin{figure}[ht]
\centering
\includegraphics[width=0.8\linewidth]{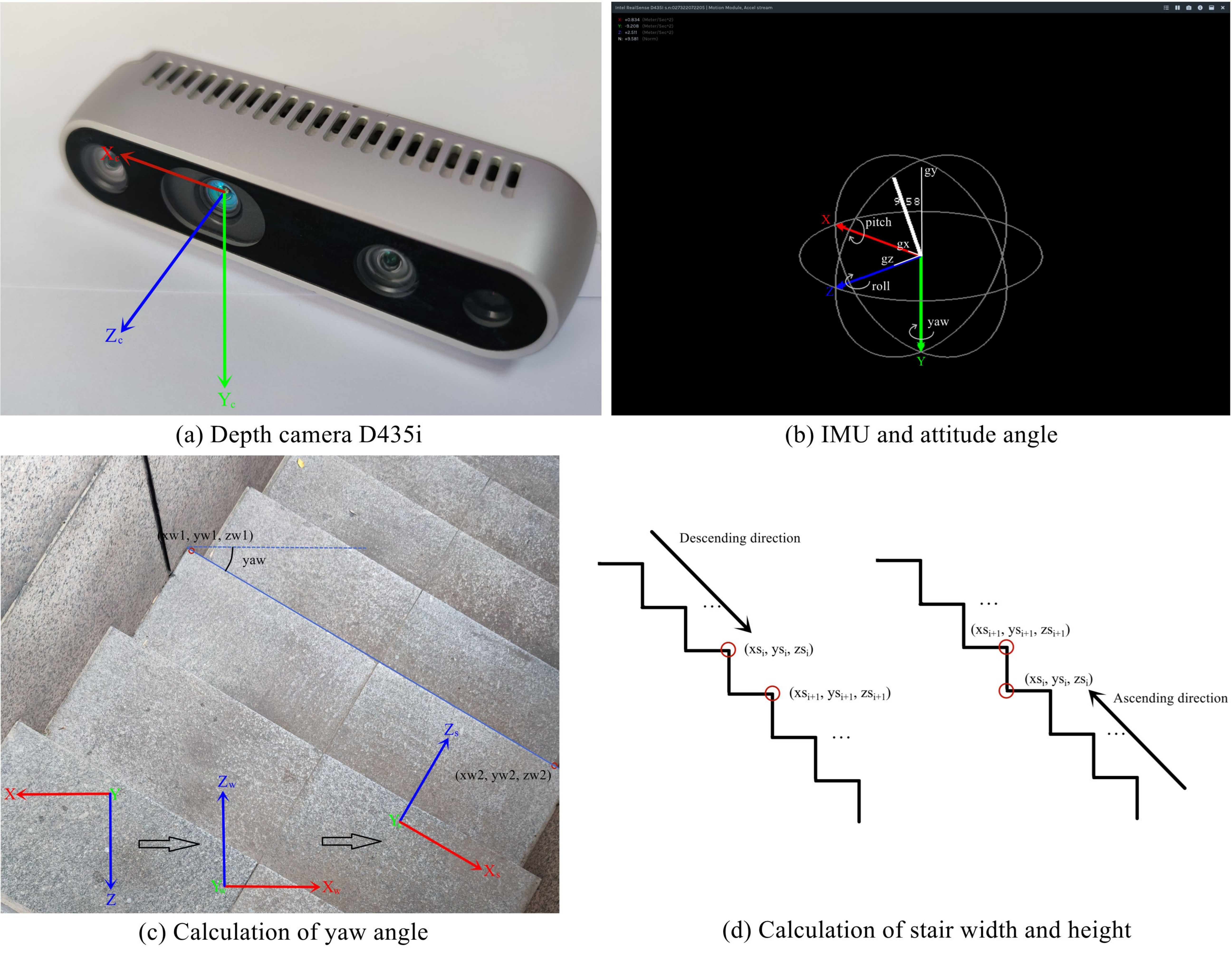}
\caption{Calculation process of stair width and height. Figure (a) shows the Realsense D435i depth camera and the definition of the camera coordinate system. Figure (b) shows the definition of the world coordinate system in the IMU module of the depth camera. When the camera axis is placed horizontally, its Z axis points to the rear of the camera. The pitch angle and roll angle can be calculated from the gravity acceleration components on the three axes, and the coordinates are transferred from the camera coordinate system to the world coordinate system. Figure (c) shows the calculation of the yaw angle, which can be calculated through any two points on a stair line in the world coordinate system. Figure (d) shows the method of calculating the width and height of stairs in the stair coordinate system $ X_{s}Y_{s}Z_{s} $.}
\label{fig5}
\end{figure}

Where, $ k_{1} $, $ k_{2} $, $ b_{1} $ and $ b_{2} $ are the parameters to be determined. x, y and z are the three-dimensional coordinates in space. After the parameters are determined, let x=0 to obtain $ z=-\frac{b_{1}}{k_{1}} $ and $ y=\frac{k_{1}b_{2} - k_{2}b_{1}}{k_{1}} $. Therefore, we can obtain the intersection point $ (0, \frac{k_{1}b_{2} - k_{2}b_{1}}{k_{1}}, -\frac{b_{1}}{k_{1}}) $ of the stair line and the camera YOZ plane. However, the intersection points are obtained in the camera coordinate system, To calculate the geometric stair parameters, we need to convert the intersection points to the world coordinate system, which requires the camera's attitude for coordinate transformation. In our research, Intel's Realsense D435i depth camera\cite{bib30} is used, and its inertial measurement unit(IMU) module can obtain the three components gx, gy and gz of gravity acceleration in the world coordinate system. In the world coordinate system defined by the D435i depth camera, the $ Y_{c} $ axis is vertically downward, the $ Z_{c} $ axis is horizontally forward and the $ X_{c} $, $ Y_{c} $ and $ Z_{c} $ axes meet the right-hand rule, as shown in Fig. \ref{fig5} (a). We use the attitude angle to describe the attitude of the camera. The attitude angle includes pitch, roll and yaw. The pitch angle is the angle between the camera $ Z_{c} $ axis and the world coordinate system XOZ, and the head up direction of the camera is the positive direction. The roll angle is the angle between the camera $ Y_{c} $ axis and the vertical plane containing the camera $ Z_{c} $ axis, and the right tilt direction of the camera is the positive direction. The yaw angle is always 0 because the Z axis defined by the world coordinate system is in the same vertical plane as the camera $ Z_{c} $ axis. The pitch angle and roll angle can be calculated using equation \ref{eq8}.

\begin{equation}
	\begin{cases}
	
	roll=arcsin(\frac{-gx}{\sqrt{gx^2+gy^2+gz^2}})\\
	pitch=arcsin(\frac{gz}{\sqrt{gx^2+gy^2+gz^2}})
	
	\end{cases}\label{eq8}
\end{equation}

The above definition of the world coordinate system is built for the IMU of D435i depth camera, and its Z axis points to the rear of the camera. For the convenience of calculation, as shown in Fig. \ref{fig5} (b), we redefine the world coordinate system $ X_{w}Y_{w}Z_{w} $, which is obtained by rotating the original world coordinate system 180 ° around the Y axis, namely, it has a reverse Y axis and Z axis compared with the original world coordinate system. After obtaining the pitch angle and roll angle, the coordinates (xc, yc, zc) in the camera coordinate system can be transferred to the coordinates (xw, yw, zw) in the world coordinate system through the coordinate transformation formula. Finally, we need to transfer the coordinates (xw, yw, zw) in the world coordinate system to the coordinates (xs, ys, zs) in the stair coordinate system. According to the architectural design principles, the $ Y_{s} $ axis and $ X_{s}OZ_{s} $ plane of the stair coordinate system should coincide with the world coordinate system. Namely, only the angle between the $ Z_{s} $ axis of the stair coordinate system and the $ Z_{w} $ axis of the world coordinate system is needed. This angle can also be understood as the yaw angle of the camera relative to the stair, as shown in Fig. \ref{fig5} (c). We use two sampling points on the same stair line to obtain the yaw angle. To improve the accuracy, we can use different sampling points to calculate the yaw angles and then calculate the mean value.

After obtaining the coordinate (xs, ys, zs) in the stair coordinate system, we can obtain the cross section of the stair. As shown in Fig. \ref{fig5} (d), we can distinguish whether to go up or down the stair using very simple prior knowledge; that is, when going up the stairs, we can see convex and concave lines at the same time, but when going down the stairs, we only see convex lines. Whether going up or down the stairs, we use equation \ref{eq9} to calculate the width and height of the stairs:

\begin{equation}
	\begin{cases}
	
	height=|ys_{i+1}-ys_{i}|\\
	width=|zs_{i+1}-zs_{i}|
	
	\end{cases}\label{eq9}
\end{equation}

Where, i represents the i-th stair line detected, and the value range is [0, the total number of detected stair lines -1]. i+1 represents the stair line closest to the i-th stair line on the side away from the camera. For the case of the ascending direction, since concave lines and convex lines can be detected at the same time, the calculated height or width of the stair lines on the same horizontal or vertical plane will be close to zero. Therefore, we filter the calculation results of equation \ref{eq9}. We discard values less than a threshold value $ \omega $ and $ \omega $= 0.05m in our study.

\section*{Experiment results}

\subsection*{Experimental settings}\label{subsec5_1}

\subsubsection*{Dataset introduction}
 
The images in the dataset are all from actual scenes. We use a Realsense D435i depth camera to capture RGB images and aligned depth images in several actual scenes. These images are segmented, padded, and scaled to 512 × 512. The whole dataset has a total of 5992 images, which are randomly divided into 4776 images for the training set and 1216 images for the validation set.

The annotation form of the dataset is as follows:

cls x1 y1 x2 y2 /n

...

Each stair line is represented by the above five-tuple data, where cls represents the class of the stair line, 0 represents a convex line, and 1 represents a concave line; x1 and y1 represent the pixel coordinates of the left endpoint of the stair line, and x2 and y2 represent the pixel coordinates of the right endpoint of the stair line. In addition, although StairNetV2 does not classify stair lines, to ensure the completeness of annotation information, we still label stair lines according to two classifications.

\subsubsection*{Training strategy}

We train the model on a workstation with an R9 5950X CPU and an RTXA4000 GPU using the PyTorch framework. As mentioned above, the input size of the network is 512 $\times$ 512, training is conducted for a total of 200 epochs, and the batch size is set to 4. We use the Adam optimizer, the weight decay is set to $ 10^{-6} $, the initial learning rate is set to 0.001, and the learning rate is halved every 50 epochs. During training, we apply the loss function with dynamic weights introduced in section \href{subsec4_2}{"Loss function with dynamic weights"}. After each epoch, the weights are dynamically adjusted according to the evaluation.

In terms of data enhancement, we use a random mirror with a probability of 0.5 to eliminate the uneven distribution of ROIs in the training images.

\subsubsection*{Evaluation metrics}

We basically follow the evaluation method of StairNet, using the accuracy, recall and intersection over union(IOU)\cite{bib31} when the confidence is 0.5 as the evaluation index. The confidence calculation is the same as that of StairNet. The difference is that StairNetV2 does not determine the classification of stair lines, and we do not focus on the background class. Therefore, the frequency weighted intersection over union(FWIOU)\cite{bib31} of StairNet can be simplified to only calculate the IOU of the class containing stair lines. The formulas of accuracy, recall and IOU are as follows:

\begin{equation}
	\begin{cases}
	
	accuracy=\frac{TP}{TP+FP}\\
	recall=\frac{TP}{TP+FN}\\
        IOU=\frac{TP}{TP+FP+FN}
	
	\end{cases}\label{eq10}
\end{equation}

Where TP, FP and FN represent the number of true positives, false positives and false negatives, respectively,  when the confidence is 0.5. Specifically, TP represents the number of cells predicted by the network to include stair lines, which actually include stair lines. FP represents the number of cells predicted by the network to include stair lines but do not actually include stair lines. FN represents the number of cells predicted by the network to contain no stair lines but actually contain stair lines.

For the final evaluation of the geometric stair parameters, considering factors such as the moving speed of the equipment, the measurement error of the depth camera and the sparsity of the remote point cloud, we only evaluate the geometric parameters of the three stair steps closest to the camera. We use the absolute error and relative error between the measurement results and the actual geometric sizes of the stairs for evaluation. In addition, to show the adaptability of the algorithm in different scenes, we select some stairs with different building structures and textures for evaluation.

\subsection*{Ablation experiments}

In this section, we conduct several ablation studies for the selective module and loss function with dynamic weights. The experiments are all conducted with the same settings as those described above. Since the selective module belongs to the category of network architecture and loss function with dynamic weights belongs to the category of network training process, we study them separately. First, we try to put the selective module into different network positions, which are after the second downsampling, after the third downsampling and after the fourth downsampling. The results are shown in  Table \ref{tab2}.

\begin{table}[ht]
\centering
\begin{tabular}{|l|l|l|l|}
\hline
Selective module position & Accuracy ($\%$)	& Recall ($\%$) & IOU ($\%$)  \\
\hline
After the second downsampling & 91.99 & 93.15 & 86.16  \\
\hline
After the third downsampling & 91.52 & 92.96 & 85.59  \\
\hline
After the fourth downsampling & 90.35 & 92.37 & 84.08  \\
\hline
\end{tabular}
\caption{\label{tab2}Ablation experiment results of selective module.}
\end{table}

It can be seen in Table \ref{tab2} that when the selective module is located after the second downsampling, the network has the best performance, which indicates that it is unreasonable to simply place a concatenate structure similar such as the selective module in a certain position of the network (such as the input and output). For StairNetV2, the network can obtain better performance by extracting the features of the two branches in the shallow layer (bottleneck1.0 - 1.2) and then further extracting the features of the fused information in the deep layer (bottleneck2.0 - 3.7).

After the selective module position is determined, we use the loss function with fixed weights and the loss function with dynamic weights to conduct a loss function ablation experiment. The results are shown in Table \ref{tab3}.

\begin{table}[ht]
\centering
\begin{tabular}{|l|l|l|l|l|}
\hline
Backbone & Loss function & Accuracy ($\%$)	& Recall ($\%$) & IOU ($\%$)  \\
\hline
StairNetV2 & Fixed weights & 91.79 & 93.15 & 85.98  \\
\hline
StairNetV2 & Dynamic weights & 91.99 & 93.15 & 86.16  \\
\hline
\end{tabular}
\caption{\label{tab3}Ablation experiment results of loss function.}
\end{table}

It can be seen from Table \ref{tab3} that using the loss function with dynamic weights to train the network can improve the network performance, which shows that the reasonable adjustments of some hyperparameters in the network training process can not only accelerate network convergence but also improve model performance.

\subsection*{Performance experiments}

In this section, we present the performance experiments of the model on the validation set, including the size, inference speed and accuracy of the model. Similar to StairNet, we provide three versions of StairNetV2, including StairNetV2 1 $\times$, StairNetV2 0.5 $\times$ And StairNetV2 0.25 $\times$, to meet the requirements of hardware devices with different computing capabilities. The size of the model is adjusted using a channel width factor that is directly multiplied by the number of channels in the network. We test the three models on desktop and mobile platforms. The specific experimental platforms, model sizes and inference speeds are shown in Table \ref{tab4}.

\begin{table}[ht]
\centering
\begin{tabular}{|l|l|l|l|l|}
\hline
Platform & StairNetV2 1 $\times$ (10.4MB) & StairNetV2 0.5 $\times$ (2.87MB) & StairNetV2 0.25 $\times$ (0.95MB) \\
\hline
R9 5950X + RTXA4000 & 11.09ms & 5.31ms & 3.06ms \\
\hline
i7 10750H + RTX2060Max-P & 19.57ms & 9.62ms & 4.90ms \\
\hline
\end{tabular}
\caption{\label{tab4}Model inference speed experiment results.}
\end{table}

The results show that the three models of StairNetV2 can meet the real-time requirements of both desktop and mobile platforms. Compared with the three models of StairNet, our model is approximately 1/3 the size of the corresponding StairNet model, so we can obtain better real-time performance. To evaluate of the accuracy of the models, we divide the data in the validation set into a daytime dataset and a nighttime dataset according to the actual collection conditions. The experimental results are shown in Table \ref{tab5}.

\begin{table}[ht]
	\centering
	\begin{tabular}{|l|l|l|l|l|l|l|}
		\hline
		\multirow{2}{*}{Model} & \multicolumn{2}{|l|}{Accuracy(\%)} & \multicolumn{2}{|l|}{Recall(\%)} & \multicolumn{2}{|l|}{IOU(\%)} \\
		\cline{2-7}
		\multirow{2}{*}{~} & Day & Night & Day & Night & Day & Night \\
		\hline
		StairNetV2 1 $\times$ & 92.88 & 90.78 & 93.34 & 92.87 & 87.11 & 84.87  \\
        \hline
        StairNetV2 0.5 $\times$ & 92.05 & 89.22 & 93.94 & 92.91 & 86.89 & 83.53  \\
        \hline
        StairNetV2 0.25 $\times$ & 91.06 & 88.21 & 92.85 & 91.87 & 85.09 & 81.82 \\
		\hline
	\end{tabular}
\caption{\label{tab5}Results of the model accuracy experiments.}
\end{table}

\begin{figure}[ht]
\centering
\includegraphics[width=0.8\linewidth]{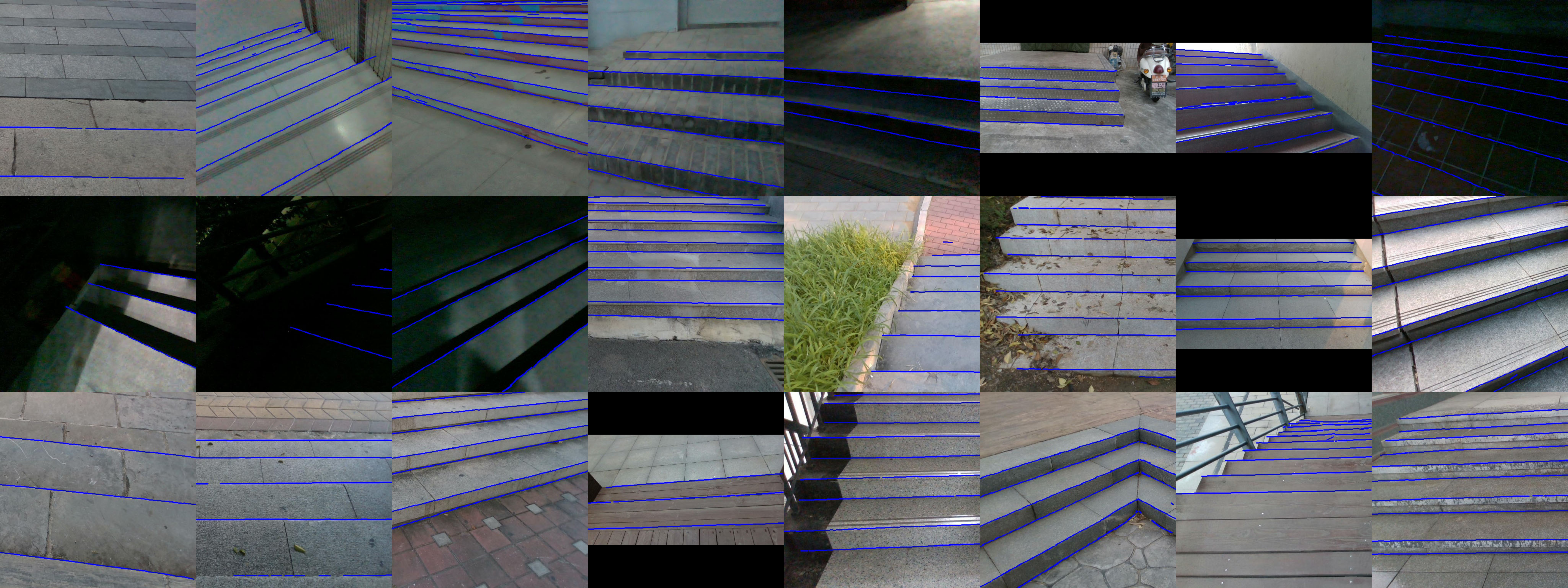}
\caption{Partial visualization results produced by StairNetV2 1 $\times$ on the validation set.}
\label{fig6}
\end{figure}

The results show that the performance of the 0.5 $\times$ model is slightly worse than that of the 1 $\times$ model, while the performance of the 0.25 $\times$ model is much lower than that of the 0.5 $\times$ and 1 $\times$ model. The 1 $\times$ model is more suitable for applications with high accuracy requirements, while the 0.25 $\times$ model is more suitable for applications with high real-time requirements, and the 0.5 $\times$ model provides a compromise option. The particularly striking result is the difference in performance between the day and night scenes. The performance of the model is very close in the day and night scenes, which is significantly improved compared with the performance of StairNet in the night scenes, and this proves the validity of the depth map for the night scenes. Fig. \ref{fig6} shows some visualization results produced by StairNetV2 1 $\times$ on the validation set. These stairs have different building materials, shooting angles and lighting conditions. It can be seen that our method still obtains pleasant results under the conditions of extreme lighting, different shooting angles, and special stair materials, especially extremely fuzzy visual cues at night.

\subsection*{Comparison experiments}

In this section, we compare our method with some existing methods, including the deep learning StairNet method, a method based on the traditional image processing algorithm and a deep learning method combined with traditional image processing. For StairNet, because the sizes of the output tensors are different from those of StairNetV2, it is necessary to adjust them to be consistent for comparison. For the traditional image processing method, we use Gabor filter, Canny edge detection and Hough transform for our implementation. For the deep learning method combined with traditional image processing, we apply YOLOv5\cite{bib28} to obtain the ROI including stair lines and then apply the traditional image processing algorithm to extract stair lines within the ROI.

Before the performance evaluation, StairNet and StairNetV2 are trained on our training set with the same settings described in section \href{subsec5_1}{"Experimental settings"}, and then we adjust the parameters of Gobor filter, Canny edge detection and Hough transform to fit our validation set. These experiments are all conducted on our validation set using the experimental platform and evaluation metrics described in section \href{subsec5_1}{"Experimental settings"}. The experimental results are shown in Table \ref{tab6}.

\begin{table}[ht]
	\centering
	\begin{tabular}{|l|l|l|l|l|}
		\hline
		Method & Accuracy (\%) & Recall (\%) & IOU (\%) & Runtime (ms) \\
        \hline
        Gabor + Canny + Hough (our implementation) & 26.02 & 25.20 & 14.68 & 7.42 \\
        \hline
        YOLOv5 + Gabor + Canny + Hough (our implementation) & 38.56 & 30.65 & 20.59 & 25.17 \\
        \hline
        StairNet 0.25 $\times$ & 83.68 & 85.59 & 73.35 & 3.71 \\
        \hline
        StairNet 0.5 $\times$ & 85.56 & 85.04 & 74.37 & 7.24 \\
        \hline
        StairNet 1 $\times$ & 86.35 & 85.18 & 75.07 & 14.90 \\
        \hline
        StairNetV2 0.25 $\times$ - Ours & 89.85 & 92.44 & 83.70 & 3.06 \\
        \hline
        StairNetV2 0.5 $\times$ - Ours & 90.86 & 93.51 & 85.46 & 5.31 \\
        \hline
        StairNetV2 1 $\times$ - Ours & 91.99 & 93.15 & 86.16 & 11.09 \\
		\hline
	\end{tabular}
\caption{\label{tab6}Comparison experiment results.}
\end{table}

\begin{figure}[ht]
\centering
\includegraphics[width=0.9\linewidth]{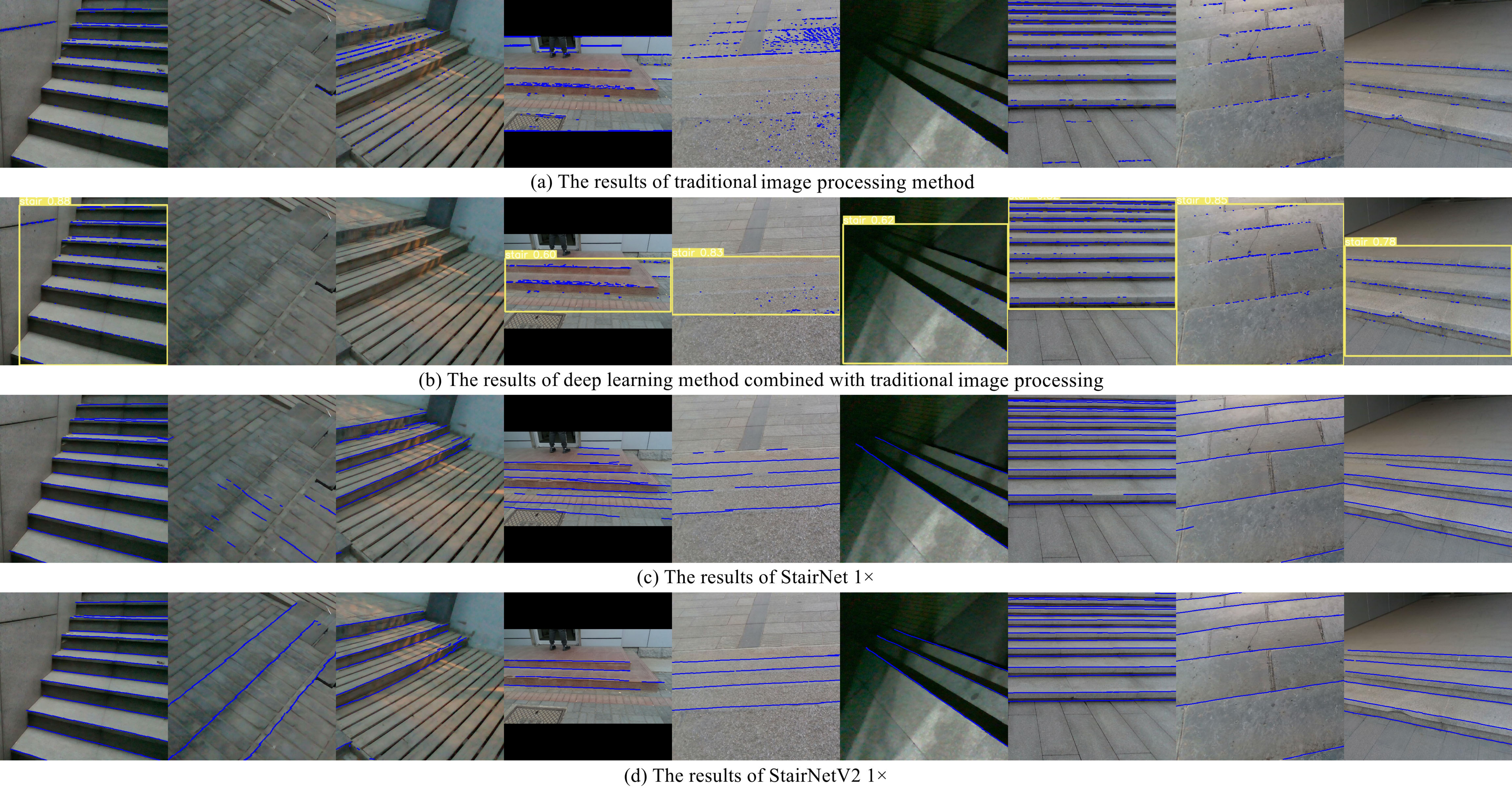}
\caption{Partial visualization results of the comparison experiments on the validation set.}
\label{fig7}
\end{figure}

The results show that compared with existing state-of-the-art depth learning method, StairNetV2 has a significant improvement in both accuracy and recall, especially the 0.25 $\times$ model of StairNetV2 still significantly surpass the 1 $\times$ model of StairNet. And compared with the 0.25 $\times$ model of StairNet, the 0.25 $\times$ model of StairNetV2 has faster speed. We benefit greatly from the input of multimodal information and the loss function with dynamic weights. In addition, compared with the methods based on traditional image processing algorithms, deep learning methods benefit from powerful learning ability and still show far more performance than traditional methods. Fig. \ref{fig7} shows partial visualization results of the above methods on our validation set. It can be seen intuitively that the performance of the methods based on deep learning are far superior to the traditional methods in various detection scenes, and StairNetV2 has better performance than StairNet in nighttime scenes and scenes with fuzzy visual clues.

\subsection*{Geometric stair parameter measurement experiments}

\begin{figure}[ht]
\centering
\includegraphics[width=1.0\linewidth]{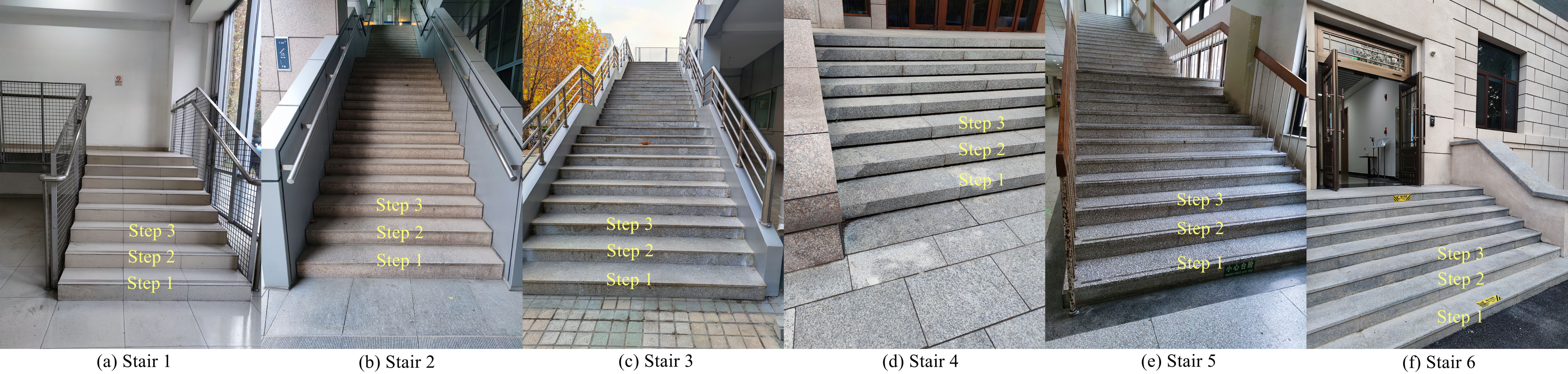}
\caption{Experimental scenes of the geometric stair parameter measurement experiments.}
\label{fig8}
\end{figure}

After obtaining the StairNetV2 output, we finally obtain the geometric stair parameters through postprocessing algorithms, including stair line clustering based on the least squares method and coordinate transformation based on the attitude angle. In this section, we measure the stairs in several actual scenes. We select several detection scenes with different building materials, different building structures and different lighting conditions for experiments, as shown in Fig. \ref{fig8}. We use the absolute error and relative error to evaluate the measurement results and separately evaluate the two walking directions of ascending and descending. The experimental results are shown in Table \ref{tab7}.

\begin{table}[ht]
\centering
\resizebox{\textwidth}{45mm}{
\begin{tabular}{|l|l|l|l|l|l|l|l|l|}
    \hline
    \multicolumn{2}{|l|}{\multirow{2}{*}{Detection scenes}} & \multirow{2}{*}{True size (mm)} & \multicolumn{2}{|l|}{Measurement size (mm)}  & \multicolumn{2}{|l|}{Absolute error (mm)}  & \multicolumn{2}{|l|}{Relative error (\%)}\\
    \cline{4-9}
    \multicolumn{2}{|l|}{~} & ~ & Ascending & Descending & Ascending & Descending & Ascending & Descending \\
    \hline
    \multirow{3}{*}{Stair1} & Step1 & (296, 160) & (296, 149) & (336, 173) & (0, -11) & (40, 13) & (0.00, -6.88) & (13.51, 8.12)\\
    \cline{2-9}
    ~ & Step2 & (293, 150) & (282, 141) & (313, 158) & (-11, -9) & (20, 8) & (-3.75, -6.00) & (6.83, 5.33) \\
    \cline{2-9}
    ~ & Step3 & (295, 150) & (277, 139) & (327, 170) & (-18, -11) & (32, 20) & (-6.10, -7.33) & (10.85,13.33) \\
    \hline
    \multirow{3}{*}{Stair2} & Step1 & (288, 153) & (290, 143) & (302, 156) & (2, -10) & (14, 3) & (0.69, -6.54) & (4.86, 1.96)\\
    \cline{2-9}
    ~ & Step2 & (289, 162) & (285, 144) & (314, 181) & (-4, -18) & (25, 19) & (-1.38, -11.11) & (8.65, 11.73) \\
    \cline{2-9}
    ~ & Step3 & (289, 161) & (282, 141) & (321, 160) & (-7, -20) & (32, -1) & (-2.42, -12.42) & (11.07, -0.62) \\
    \hline
    \multirow{3}{*}{Stair3} & Step1 & (304, 153) & (304, 137) & (315, 155) & (0, -16) & (11, 2) & (0.00, -10.46) & (3.62, 1.31)\\
    \cline{2-9}
    ~ & Step2 & (307, 145) & (304, 134) & (329, 163) & (-3, -11) & (22, 18) & (-0.98, -7.59) & (7.17, 12.41)\\
    \cline{2-9}
    ~ & Step3 & (307, 145) & (302, 129) & (328, 157) & (-5, -16) & (21, 12) & (-1.63, -11.03) & (6.84, 8.28)\\
    \hline
    \multirow{3}{*}{Stair4} & Step1 & (341, 143) & (346, 126) & (372, 152) & (5, -17) & (31, 9) & (1.47, -11.89) & (9.09, 6.29)\\
    \cline{2-9}
    ~ & Step2 & (340, 147) & (342, 127) & (372, 145) & (2, -20) & (32, -2) & (0.59, -13.61) & (9.41, -1.36)\\
    \cline{2-9}
    ~ & Step3 & (342, 146) & (340, 131) & (367, 154) & (-2, -15) & (25, 8) & (-0.58, -10.27) & (7.31, 5.48)\\
    \hline
    \multirow{3}{*}{Stair5} & Step1 & (300, 163) & (301, 150) & (321, 159) & (1, -13) & (21, -4) & (0.33, -7.98) & (7.00, -2.45)\\
    \cline{2-9}
    ~ & Step2 & (298, 154) & (303, 139) & (313, 162) & (5, -15) & (15, 8) & (1.68, -9.74) & (5.03, 5.19)\\
    \cline{2-9}
    ~ & Step3 & (296, 154) & (288, 144) & (322, 175) & (-8, -10) & (26, 21) & (-2.70, -6.49) & (8.78, 13.64) \\
    \hline
    \multirow{3}{*}{Stair6} & Step1 & (321, 114) & (321, 108) & (336, 127) & (0, -6) & (15, 13) & (0.00, -5.26) & (4.67, 11.40)\\
    \cline{2-9}
    ~ & Step2 & (322, 151) & (326, 137) & (332, 148) & (4, -14) & (10, -3) & (1.24, -9.27) & (3.11, -1.99)\\
    \cline{2-9}
    ~ & Step3 & (322, 152) & (327, 137) & (352, 156) & (5, -15) & (30, 4) & (1.55, -9.87) & (9.32, 2.63)\\
    \hline
    \multicolumn{2}{|l|}{Root mean square} &  &  &  & (6.29, 14.23) & (24.82, 11.40) & (2.11, 9.40) & (8.07, 7.71) \\
    \hline
\end{tabular}}
\caption{\label{tab7}Results of the geometric stair parameter measurement experiments. A cell is denoted as (width, height).}
\end{table}

The results show that for the case of ascending stairs, the measurements of stair width are accurate, the measurement errors of height are relatively large, and the measured heights are less than the true values in most cases. For the case of descending stairs, the measurement values are not stable. In most cases, the measurements values exceed the true values to some extent. The main error sources of the whole process are as follows: 1) The error caused by the jitter of the detection algorithm. 2) The error of point cloud data obtained by the depth camera, including the error of distance measurement, the black hole of point cloud at the edges of the stairs and at the reflective areas of smooth stairs. 3) The error of the IMU leads to the error of the obtained angles,which affects the coordinate transformation.

\section*{Conclusion}

We propose a novel neural network architecture with multimodal input for stair detection and implement a complete RGB-D-based deep learning stair detection method in an end-to-end manner. In this way, We overcome the problem of reliable detection at night and in the case of extremely fuzzy visual clues. We also propose a loss function with dynamic weights that can optimize the network training process and several postprocessing algorithms for obtaining the geometric stair parameters. In addition, we provide an RGB-D stair dataset with fine annotations for stair detection research. Experiments conducted on the dataset demonstrate the effectiveness and performance of our method. The root mean square errors of the geometric stair parameters obtained with postprocessing algorithms are within 15mm when ascending stairs and 25mm when descending stairs. In future work, the point cloud data will be sent to a neural network to obtain the geometric stair parameters  more efficiently.

\section*{Data availability}

Our dataset is available at  \href{https://data.mendeley.com/datasets/p28ncjnvgk}{https://data.mendeley.com/datasets/p28ncjnvgk}.

\bibliography{main.bib}

\section*{Author contributions statement}

C.W made the dataset. C.W and S.Q made all the figures used in the paper. C.W designed the software architecture and wrote the paper. C.W and Z.T conceived the experiments and conducted the experiments. Z.T, S.Q and Z.P analysed the results. All authors reviewed the manuscript.

\section*{Competing interests}

The authors declare no competing interests.

\end{document}